# SAR SHIP DETECTION BASED ON SWIN TRANSFORMER AND FEATURE ENHANCEMENT FEATURE PYRAMID NETWORK

*Xiao Ke, Xiaoling Zhang, Tianwen Zhang, Jun Shi, Shunjun Wei*

School of Information and Communication Engineering
University of Electronic Science and Technology of China, Chengdu, China
*Corresponding author. E-mail: xke@std.uestc.edu.cn; xlzhang@uestc.edu.cn;
twzhang@std.uestc.edu.cn; shijun@uestc.edu.cn; weishunjun@uestc.edu.cn

**ABSTRACT**

With the booming of Convolutional Neural Networks (CNNs), CNNs such as VGG-16 and ResNet-50 widely serve as backbone in SAR ship detection. However, CNN based backbone is hard to model long-range dependencies, and causes the lack of enough high-quality semantic information in feature maps of shallow layers, which leads to poor detection performance in complicated background and small-sized ships cases. To address these problems, we propose a SAR ship detection method based on Swin Transformer and Feature Enhancement Feature Pyramid Network (FEFPN). Swin Transformer serves as backbone to model long-range dependencies and generates hierarchical features maps. FEFPN is proposed to further improve the quality of feature maps by gradually enhencing the semantic information of feature maps at all levels, especially feature maps in shallow layers. Experiments conducted on SAR ship detection dataset (SSDD) reveal the advantage of our proposed methods.

## 1. INTRODUCTION

SAR ship detection plays an important role in many fields such as disaster rescue, marine management and boarder monitoring. Traditional SAR ship detection methods can be implemented to specific cases with modest performance, but need different sea clutter distribution model and appropriate parameters for different SAR images, which is hard to achieve high precision detection in unknown SAR images. With the breakthrough of deep learning in pedestrian detection, face recognition and other vision tasks in recent years, SAR ship detection algorithms based on deep learning are gradually emerging.

Kang [1] fused deep semantic and high-resolution features from feature maps of different layers to enhance the detection performance for small-sized ships. Lin [2] used squeeze and excitation mechanism to further improve feature quality of multiscale feature maps. Zhao [3] proposed attention receptive pyramid network to detect multiscale ships. Wei [4] proposed a high-resolution ship detection network to fully utilize the feature maps of high-resolution and low-resolution convolutions. All the methods above used CNN based backbone to acquire abstract feature maps and then fuse these features maps to generate more discriminative features. However, due to the limited receptive field of single convolution, features with long-range dependencies are difficult to acquire even repeating single convolution many times [5], and feature maps in shallow layers with fewer repeated convolutions operations are short of high-quality semantic information, which leads to the difficulties in detecting small-sized ships especially in complicated background. Even though Lin proposed FPN [6] that adds a top-down path to propagate high-level semantic feature at all scales, it may still not enough for SAR ship detection mission where the number of small-sized ships occupy a large proportion and feature maps in shallow layers generate the majority of ship proposal.

In this paper, we propose a SAR ship detection method based on Swin Transformer [7] and Feature Enhancement Feature Pyramid Netwrok(FEFPN). Swin Transformer is introduced as backbone to capture long-range dependencies and overcome the defect of convolution. FEFPN is proposed to gradually strengthen the semantic information in feature maps of shallow layers based on two prior knowledge about Swin Transformer and SAR ship detection. Ablation experiments and comparative experiments show the effectiveness of our methods.

## 2. METHODOLOGY

Our method follows the pipeline of two-stage detectors featuring high accuracy and modest cost. Figure 1 shows the architecture of our proposed SAR ship detection method. Swin-T (tiny version of Swin Transformer) serves as backbone to extract hierarchical feature maps for the following feature pyramid network. Notably, we do not use larger versions such as Swin-S or Swin-B since the complexity of Swin-T is similar to that of widely used ResNet-50. After feature extraction, hierarchical feature maps serve as input of our proposed Feature Enhancement Pyramid Network to enhance the feature of themselves. We keep the same region proposal network and classification and bounding box regression network (denoted as RPN and Fast-RCNN respectively in figure 1) as Faster R-CNN [11]. More details are described in 2.1 and 2.2.

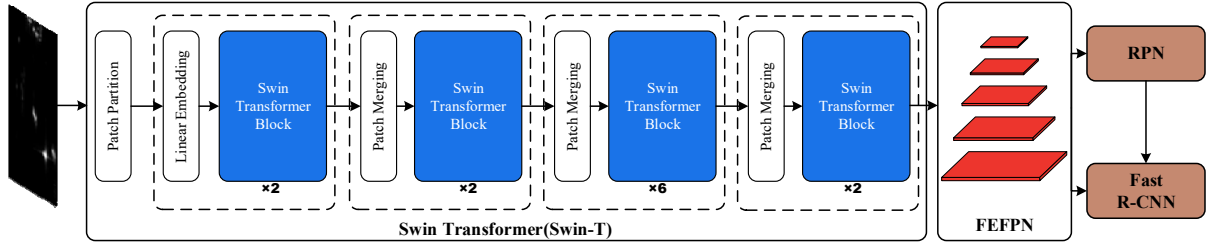

Fig. 1. The architecture of our proposed SAR ship detection method

### 2.1. Swin Transformer

Swin Transformer first splits a SAR image into non-overlapping patches. Each patch is treated as a "token" and is projected to an arbitrary dimension (denoted as C and C = 96 in Swin-T) by linear embedding. Then two successive Swin Transformer blocks are applied on these patch tokens and generate the first stage of feature maps. The second, third and fourth stages of feature maps are also generated by similar successive Swin Transformer blocks respectively, except for the number of used blocks is different in different stages (2, 2, 6, 2 in the first, second, third, fourth stage respectively). Notably, to produce hierarchical feature maps with the same feature map resolutions as ResNet, patch merging layer is applied between every two stages and is used to conduct 2× downsampling of resolution and double the output dimension.

Figure 2 shows the architecture of two successive Swim Transformer Blocks. Swin Transformer Block is built by replacing the standard multi-head self attention (MSA) module in a Transformer Block by a module based on shifted windows (i.e. W-MSA and SW-MSA in figure 2), with a LayerNorm (LN) layer applied before each MSA module and a two-layer MLP deployed in the end of Swim Transformer Block. Different from MSA, W-MSA computes self-attention in local non-overlapping windows instead of conducting global self-attention. This window-based self-attention makes Swin Transformer achieve linear computational complexity, while standard Transformer has quadratic complexity with respect to image size. However, window-based self-attention is short of cross-window connections, which may decrease the modeling power. Therefore, SW-MSA, based on a shifted window configuration from MSA's window configuration, is used to enhance cross-window connections.

There are three reasons why we choose Swin Transformer as backbone instead of CNN based backbone such as ResNet-50. First, and most importantly, Transformer-like architectures are notable for modeling long-range depend-encies and this advantage is helpful for SAR ship detection especially in complex background occasions. Secondly, Swin Transformer only has linear complexity in comparison with other vision Transformer such as [8] which has quadratic complexity. Thirdly, Transformer-like architectures derive from the dominant architecture (i.e. Transformer) in NLP, hence has much potential in future studies such as multi-modal SAR ship detection due to the unified modeling ability.

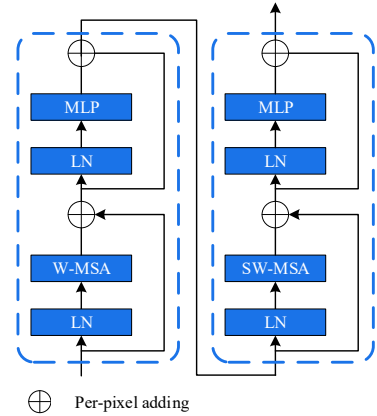

Fig. 2. Two successive Swin Transformer Blocks

### 2.2. FEFPN

Figure 3 shows the architecture of our proposed Feature Enhancement Feature Pyramid Network (FEFPN). FEFPN consists of three top-down paths with different weights, several lateral connections and three residual connections. The first top-down path is totally the same as the original FPN [6]. In the feature fusing part of first top-down path (as shown in the left rectangular box of figure 3), 1×1 convolution is used for channel matching (from (96, 192, 384, 768) to (256, 256, 256, 256)); 2× up sampling is used for resolution matching; 3×3 convolution is used to reduce the aliasing effect of upsampling after per-pixel adding. The second top-down path is similar to the first one except for two differences which can be told form the middle rectangular box in figure 3. One difference is that we did not add 1×1 convolution for channel matching because the channel of adjacent feature maps in the second top-down path has already been the same. The other difference is that before per-pixel adding, we add a per-feature weighting part whose weight is decayed to $0.5\times$ because we regard the feature enhancement progress as a finetuning progress. The third top-down path is basically the same as the second one except for one difference that the weight of per-feature weighting part is further decayed to $0.25\times$ (as shown in the right rectangular box in figure 3). Additionally, to accelerating the converging of our network, we add three residual connections from three shallow layers in the first top-down path to the corresponding layers in the third top-down path, which is shown as black curves in figure 3.

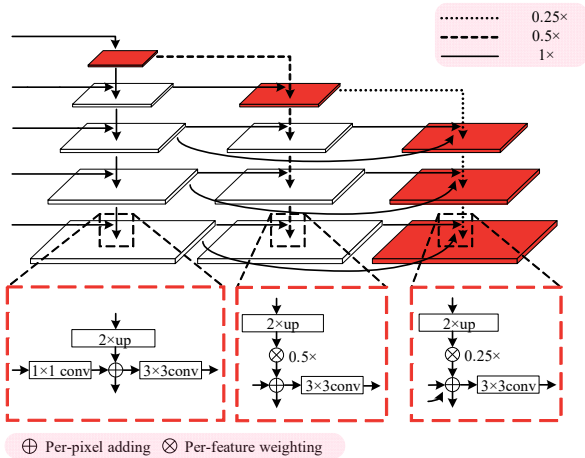

Fig. 3. Feature Enhancement Feature Pyramid Network

FEFPN is designed to enhance the feature quality in shallow layers and origins from two prior knowledge. One prior knowledge is that Swin Transformer is translational semi-invariance [7] while CNN is translational full-invariance, which means the quality of semantics abstracted by Swin Transformer may be better than that of geometric details abstracted by Swin Transformer, and this hypothesis can be proved to a certain extent by the ablation experiments we made in Chapter 3. Therefore, we use the abstract semantical information to enhance the feature quality by using three top-down paths which propagate semantics, instead of down-top paths which focus on the enhancement of geometric details.

The other prior knowledge is that small-sized ships dominate the dataset [9], which makes training samples of deep layers not sufficient and the feature quality of shallow layers needs more attention. Therefore, we try not to increase much complexity in feature maps of deep layers, and use three lateral connections and top-down paths with different gradually decayed weights to gradually enhance the semantics in shallow layers. Notably, the reason we choose different gradually decayed weights for three top-down path is that we regard the feature enhancement progress as a fine-tuning progress. Hence the change of feature map at the same level gradually becomes smaller as the weight of each top-down path decays. We carefully choose 0.5 and 0.25 as decayed weights because 0.5 and 0.25 proved to superior while fusing adjacent layers of FPN for tiny object through brute force search in [10].

Prior knowledge guided us in designing the architecture of FEFPN and we believe adding more prior knowledge into architecture designing SAR ship detectors is quite essential, since the volume of SAR ship detection dataset is usually smaller than that of other detection missions datasets such as COCO that is large enough to learn prior knowledge implicitly.

## 3. EXPERIMENTS

In order to prove the effectiveness of our proposed methods, we conduct ablation experiments and comparative experiments based on Pytorch, with a PC equipped with NVIDIA GeForce RTX 2070 SUPER, Intel(R) Core (TM) i7-10700 CPU @ 2.90GHz and 16G memory. We choose SSDD [9] for it has been widely used by researchers in SAR ship detection area. The proportion of training set and test set is set as 8:2, following the suggestion of [9]. RestNet-50 and Swin-T are initialized by their respective pre-training model. We follow the same loss function as Faster RCNN and used AdamW optimizer (initial learning rate of 0.0001, weight decay of 0.05, and batch size of 2 due to limited memory).

### 3.1. Evaluation indicators

We use Mean Square Precision (mAP) to evaluate performances of SAR ship detection methods. mAP is the average AP of all object categories. And mAP is the same as AP in SAR detection mission for there is only one object category. AP is defined in Equation (1), where $P$ denotes *Precision* defined in Equation (2), and $R$ denotes *Recall* defined in Equation (3).

$$AP = \int_0^1 P(R)dR \quad (1)$$

$$Precison = \frac{True\ Positives}{All\ Detections} \quad (2)$$

$$Recall = \frac{True\ Positives}{All\ Ground\ Truths} \quad (3)$$

### 3.2. Ablation Experiments

We conduct ablation experiments to prove the effectiveness of Swin-T and FEFPN. Table 1 shows the detection results of detectors using different backbones and necks, while other parts keeping the same as Faster RCNN. From Table 1, Swin-T improve the detection accuracy by 2.21% mAP due to the excellence long-range dependencies modeling ability of Swin Transformer. However, when FPN is replaced with PAFAN that features bringing more spatial details to feature maps at all levels, mAP decreases by 3.91%. This sharp decrease may derive from the fact that Swin Transformer is translational semi-invariance. Hence the quality of spatial details captured by Swin Transformer is not good enough to strength feature maps. This phenomenon enlighten us to some extent and we propose FEFPN by carefully enhancing features in the original FPN with abstract semantical information. It proves to be helpful since mAP rises from 92.51% to 93.08% by replacing FPN with FEFPN.

Table 1 Ablation Experiments

| Backbone | Neck | mAP(%) |
|---|---|---|
| ResNet-50 | FPN | 90.30 |
| Swin-T | FPN | 92.51 |
| Swin-T | PAFPN | 88.60 |
| Swin-T | FEFPN | 93.08 |

### 3.3. Comparative Experiments

Table 2 shows the comparison with other four competitive state-of-the-art detectors, all of which use CNN based backbone. From Table 2, one can see that our method outperform second-place DCN whose backbone is the improved variant of CNN. Moreover, RetinaNet, a classic one-stage detector, performs poorly compared with all the two-stage detectors such as Faster RCNN, DCN, PANET and our method.

Interestingly, by associating Table 1 and Table 2 to analyze, it can be observed that our method without FEFPN behaves almost the same as DCN (92.51% mAP versus 92.60% mAP). This finding shows that CNN based backbone and Swin Transformer based backbone seem to have its own advantage and disadvantage, with similar performance in feature abstracting. However, we highlight the strong potential of Swin Transformer in the future studies such as multi-modal SAR ship detection, since Transformer-like architectures derive from NLP and have the ability of unifying vision tasks and language tasks.

Table 2 Comparative Experiments

| Method | mAP(%) |
|---|---|
| Faster RCNN [11] | 90.30 |
| RetinaNet [12] | 87.10 |
| DCN [13] | 92.60 |
| PANET [14] | 92.30 |
| Our method | 93.08 |

### 4. CONCLUSION

We propose a SAR ship detection method based on Swin Transformer and Feature Enhancement Feature Pyramid(FEFPN). Swin Transformer is used to model long-range dependencies that CNN based backbone is not good at modeling. FEFPN is proposed to enhance the feature quality of the original FPN by carefully propagating semantics to features maps, especially to feature maps of shallow layers. Additionally, we give three reasons why we use Swin Transformer, and two prior knowledge we take into consideration while designing FEFPN. In comparison with other state-of-the-art detectors, our method achieve the best result in SSDD, with mAP of 93.08%. And the ablation experiments show the effectiveness of Swin Transformer and our proposed FEFPN.